# Activation function optimization method: Learnable series linear units (LSLUs)

Chuan Feng[1], Xi Lin[1], Shiping Zhu[*,1,2,4], Hongkang Shi[1,3], Maojie Tang[1], Hua Huang[1]

1. College of Engineering and Technology, Southwest University, Chongqing, 400716, China.
2. Yibin Academy of Southwest University, Yibin, 644000, Sichuan, China.
3. Sericulture research institute of Sichuan Academy of Agricultural Sciences, Nanchong, Sichuan, 637000, China.
4. State Key Laboratory of Resource Insects, Southwest University, Chongqing, 400700, China.

**Abstract:** Effective activation functions introduce non-linear transformations, providing neural networks with stronger fitting capabilities, which help them better adapt to real data distributions. Huawei Noah's Lab believes that dynamic activation functions are more suitable than static activation functions for enhancing the non-linear capabilities of neural networks. Tsinghua University's related research also suggests using dynamically adjusted activation functions. Building on the ideas of using fine-tuned activation functions from Tsinghua University and Huawei Noah's Lab, we propose a series-based learnable activation function called LSLU (Learnable Series Linear Units). This method simplifies deep learning networks while improving accuracy. This method introduces learnable parameters $\theta$ and $\omega$ to control the activation function, adapting it to the current layer's training stage and improving the model's generalization. The principle is to increase non-linearity in each activation layer, boosting the network's overall non-linearity. We evaluate LSLU's performance on CIFAR10, CIFAR100, and specific task datasets (e.g., Silkworm), validating its effectiveness. The convergence behavior of the learnable parameters $\theta$ and $\omega$, as well as their effects on generalization, are analyzed. Our empirical results show that LSLU enhances the generalization ability of the original model in various tasks while speeding up training. In VanillaNet training, parameter $\theta$ initially decreases, then increases before stabilizing, while $\omega$ shows an opposite trend. Ultimately, LSLU achieves a 3.17% accuracy improvement on CIFAR100 for VanillaNet (Table 3). Codes are available at https://github.com/vontran2021/Learnable-series-linear-units-LSLU.

## 1 Introduction

Regarding computer vision tasks in deep learning, the activation function plays a crucial role in neural networks. Tasks such as classification, detection, and segmentation rely on deep neural networks to extract image features[1,2]. As a key component in neural networks, the activation function performs a non-linear transformation on the output of each layer, allowing the network to learn and express complex non-linear relationships, thereby capturing more complex and abstract features[3]. Taking image feature extraction as an example, through the role of the activation function, neural networks can learn deeper-layer features such as edges, textures, and shapes, thereby enhancing the network's representation capability of the data[4].

* Corresponding author.
E-mail address: zspswu@126.com (Shiping Zhu)

Effective activation functions (such as ReLU[5], Sigmoid[6], Tanh, etc.) introduce different forms of non-linear transformations, endowing neural networks with stronger fitting capabilities, and enabling them to better adapt to real data distributions. These activation functions assist the network in converging faster during the training process, alleviate the vanishing gradient problem, and provide the network with stronger representational capabilities, thereby improving the model's performance and generalization ability. Based on this, after extensive experimental research, we propose a learnable activation function based on series, called LSLU (Learnable Series Linear Units), which can make various deep learning networks more concise and achieve higher accuracy.

## 2 Related Work

**Static activation functions.** Early activation functions such as Sigmoid and Tanh map inputs to specific ranges (e.g., Sigmoid: 0→1). However, these types of activation functions suffer from the gradient saturation problem, leading to issues like vanishing or exploding gradients during training. ReLU (Rectified Linear Unit) and its variants improve training stability by addressing this problem[7]. Their principle involves making the activation function sparse and suppressing negative inputs to mitigate the vanishing gradient problem. However, these activation functions truncate a significant amount of image information due to the cutoff of negative inputs, where gradients for negative inputs are completely set to zero during backward propagation. Leaky ReLU[8] and ELU (Exponential Linear Unit)[9] resolve the gradient truncation issue in the negative region but come with a trade-off of reduced computation speed. SELU (Scaled Exponential Linear Unit)[10] improves upon ELU by introducing a self-normalization mechanism, allowing the hidden layers of neural networks to automatically maintain output mean and variance close to 1 during training. However, SELU has strict requirements on initialization parameters and network output distribution, and improper usage may lead to unstable output mean and variance, increasing training instability. To address the challenge of retaining some negative inputs while maintaining stability, GELU (Gaussian Error Linear Unit)[11] and SiLU (Sigmoid Linear Unit)[6, 12] were introduced. These activation functions behave similarly to ReLU when inputs are positive, gradually approaching zero for negative inputs. This design preserves the efficiency of ReLU while overcoming its cutoff issue.

**Dynamic activation functions.** PReLU (Parametric ReLU)[13] is an extension of Leaky ReLU, allowing the slope parameter α to be a learnable parameter, enabling the network to automatically adjust the slope. This feature makes PReLU more adaptable to different data distributions and tasks, reducing the need for manual tuning. However, due to the lack of constraints, it increases the model's parameter count and computational overhead. To address this, researchers redesigned the slope parameter α, sampling it randomly from a predefined range, resulting in RReLU (Random ReLU)[8]. However, it still does not resolve the instability caused by random parameters and the issue of negative inputs being truncated. Therefore, CELU (ELU plus)[14] was proposed, an extension of ELU that allows both negative inputs to be retained and the slope of ELU to be a learnable parameter. However, this change increases the model's complexity. Thus, Liao Z. proposed an activation function based on Fourier series. The principle is to simulate existing basic activation functions using Fourier series and involve the coefficients in training. As shown in Figure 2-5 (b) in the bottom right, the Fourier series simulates the Leaky ReLU activation function. Although this activation function enhances the network's non-linear capabilities, it still cannot avoid the instability in model training caused by the randomness in learning the parameter value N (number of terms)[15]

In this paper, we have conducted a more in-depth exploration of the value of learnable activation functions. By enhancing the parameter update approach and incorporating concepts related to series, we propose a method for a learnable series activation function。

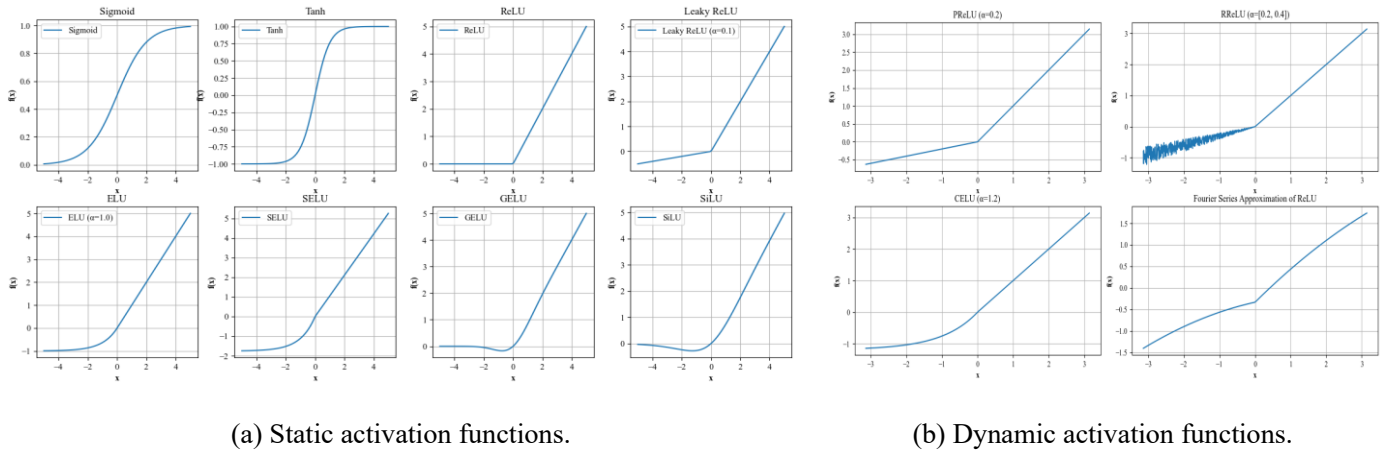

(a) Static activation functions.      (b) Dynamic activation functions.

Figure 1: Activation Functions. We define activation functions with fixed parameters as static activation functions and those with parameters that can change as dynamic activation functions.

# 3 Methods

*3.1 Background and Motivations*

The limited power of simple and shallow networks is mainly attributed to poor nonlinearity[16-18] . There are two approaches to improving the nonlinearity of neural networks: deepening the nonlinearity layers or enhancing the nonlinearity of a single layer. Currently, the mainstream approach favors the former, with limited research on enhancing the nonlinearity of a single-layer network[17]. Huawei Noah's Lab's latest research[19] reveals that increasing the nonlinearity of each activation function layer can enhance the overall network's nonlinearity, thereby significantly reducing the depth of the neural network. It is wellknown that improving the nonlinearity of activation functions can be achieved through serial stacking or parallel stacking. Serial stacking involves adding or reducing activation functions as the network deepens, while parallel stacking means simultaneously adding or reducing activation functions for all single-layer convolutions used in the network. This work focuses on the parallel stacking of activation function layers. In the forward propagation of neural networks, a single activation function $f(x)$ applied to input X, it can be a commonly used activation function such as ReLU[5]. Let the activation function $f(x)$ be represented as:

$$X_i = f(X_{i-1} + B_{i-1}) \tag{1}$$

Equation (1), where $i \in N^+$, $B_i$ represents the bias of the $i$-th layer. Then multiplied by the dynamic parameter α, we get:

$$X_i = \alpha f(X_{i-1} + B_{i-1}) \tag{2}$$

If used in parallel stacking, multiple activation functions simultaneously act on the current layer.

$$X_i = \alpha_{i-1,1} f(X_{i-1} + B_{i-1,1}) + \alpha_{i-1,2} f(X_{i-1} + B_{i-1,2}) + \cdots + \alpha_{i-1,q} f(X_{i-1} + B_{i-1,q}) + \cdots + \alpha_{i-1,n} f(X_{i-1} + B_{i-1,n}) \tag{3}$$

Expressing this stacking using a series, we have:

$$X_i = \sum_{n=1}^{n} \alpha_{i-1,n} f(X_{i-1} + B_{i-1,n}) \tag{4}$$

Let's denote the output obtained by applying $n$ activation functions to the input $X_{i-1} = \square^{C_{in} \times H_{in} \times W_{in}}$ as $X_i = \square^{C_{out} \times H_{out} \times W_{out}}$.

$$X_{i,j} = \sum_{j=1}^{C_{in}} \sum_{n=1}^{N} \alpha_{i-1,n} f(X_{i-1,j} + B_{i-1,j}) \tag{5}$$

The equation (5) involves $j$, representing the $j$-th feature matrix in the $i$-th convolutional layer, where $j$ take values $j \in \{1, 2, ..., C_{in}\}$.

In neural network architecture design, the typical pattern involves convolution followed by normalization and activation functions. For the input $x \in X_{i-1}$, the resulting feature matrix from the convolution operation is denoted as:

$$x_{out} = x_{j,c_{out},h_{out},w_{out}} = \sum_{c_{in}=1}^{C_{in}} \sum_{h_k=1}^{H_k} \sum_{w_k=1}^{W_k} x_{j,c_{in},h_{in},w_{in}} \tag{6}$$

Where, the convolutional kernel is of size $k \times k$, and the *Mean* ($\mu_{c_{out}}$) and *Variance* ($\sigma_{c_{out}}$) of the input are computed simultaneously:

$$\mu_{c_{out}} = \frac{1}{Q \times H \times W} \sum_{q=1}^{Q} \sum_{h=1}^{H} \sum_{w=1}^{W} x_{q,c,h,w} \tag{7}$$

$$\sigma_{c_{out}} = \frac{1}{Q \times H \times W} \sum_{q=1}^{Q} \sum_{h=1}^{H} \sum_{w=1}^{W} (x_{q,c,h,w} - \mu_{c_{out}})^2 \tag{8}$$

Where, $Q$ denotes the batch size of incoming images. Combining the input samples $\mu_{c_{out}}$ and $\sigma_{c_{out}}$, the output $x_{out}$ is now normalized, resulting in the output $y_{out}$:

$$y_{out} = \frac{x_{out} - \mu_{c_{out}}}{\sqrt{\sigma_{c_{out}}^2 + \varepsilon}} \cdot \gamma_{c_{out}} + \beta_{c_{out}} \tag{9}$$

Where, $\gamma_{c_{out}}$ and $\beta_{c_{out}}$ are the scaling and shifting coefficients, respectively. Both are automatically updated during back-propagation based on the training loss.

$$\gamma' = \gamma - \eta \cdot \frac{\partial L}{\partial \gamma}, \quad \beta' = \beta - \eta \cdot \frac{\partial L}{\partial \beta} \tag{10}$$

Where, $\eta$ is the learning rate, $L$ is the loss function. Finally, the non-linear activation function $\varphi(\cdot)$ is applied $y_{out}$, resulting in $X_i$.

$$y'_{out} = \varphi(y_{out}), \quad y'_{out} \in X_i \tag{11}$$

*3.2 LSLU formulation*

Hanting Chen et al.[19] modified the basic activation function to enhance the nonlinearity of a single layer by:

$$\varphi(x) = (1-\lambda)f(x) + \lambda x \tag{12}$$

Where $\lambda$ is the ratio of the current epoch to the total number of epochs, i.e., as the number of layers increases, the activation function gradually weakens with the ratio of the current epoch to the total number of epochs. When reaching the last epoch, the input and output become equal, and the activation function no longer takes effect. However, this may cause the model to learn a large amount of incorrect information, leading to instability. To mitigate the issues of shallow layers and instability, they introduced a new training strategy. Specifically, they combined the weight matrix and bias matrix with the Batch Normalization (BN) layer during the convolution and normalization stages, resulting in the fused weight matrix and bias matrix[20]:

$$W'_i = \frac{\gamma_{i-1}}{\sigma_{i-1}} W_i, \quad B'_i = \frac{(B_i - \mu_i)\gamma_i}{\sigma_i} + \beta_i \tag{13}$$

Where $i \in \{1, 2, ..., C_{out}\}$ denotes the value in *i*-th output channels, After merging the convolution with batch normalization, we begin to merge the two 1×1 convolutions. Denote $x \in \mathbb{R}^{C_{in} \times H \times W}$ and $y'_{out} \in \mathbb{R}^{C_{out} \times H_{out} \times W_{out}}$ as the input and output features, the convolution can be formulated as followed by a non-linear processing use $\varphi(\cdot)$, resulting in:

$$y'_{out} = \varphi(W * x + B) = \varphi(W \cdot im2col(x) + B) = \varphi(W \cdot X + B), \quad y'_{out} \in X_i \tag{14}$$

Where the symbol "·" denotes matrix multiplication and $X \in \mathbb{R}^{(C_{out} \times 1 \times 1) \times H_{out} \times W_{out}}$ is derived from the *im2col* operation to transform the input into a matrix corresponding to the kernel shape.

To further enhance nonlinearity and facilitate the use of learnable activation functions, we have made improvements to this method. Firstly, assuming the activation function[19] based on a series is denoted as $F(X) = \sum_{n=1}^{N} \alpha_{i-1,n} f(X_{i-1} + B_{i-1,n})$. We introduce learnable parameters $\theta$ and $\omega$ to control the oscillation amplitude and slope of the activation function, given by:

$$S(X) = \sum_{n=1}^{N} \theta_n F(X) + \omega_n \tag{15}$$

The parameters $\theta$ and $\omega$, initialized to 1 and 0, respectively, are updated based on backpropagation during training.

$$\theta' = \theta - \eta \cdot \frac{\partial L}{\partial \theta}, \quad \omega' = \omega - \eta \cdot \frac{\partial L}{\partial \omega} \tag{16}$$

At the stage of combining weight matrices and bias matrices, the weight matrix is further adjusted based on the parameter $\theta$.

$$y'_{out} = S(\theta W \cdot X + B), \quad y'_{out} \in X_i \tag{17}$$

In Figure 2 illustrates this oscillating variation.

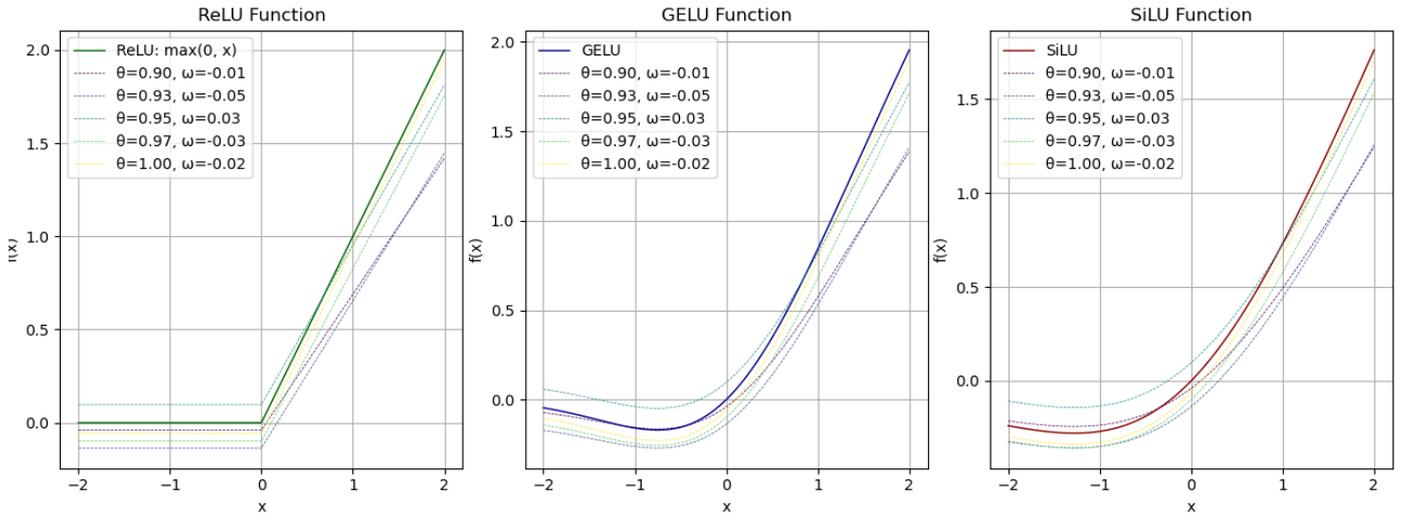

Figure 2: Changes in LSLU

To validate the capabilities of LSLU, we conducted experimental evaluations. In the next section, its performance on public datasets as well as instance application data will be demonstrated.

## 4 Experiments

In this section, we evaluated the performance of Learnable Series Linear Units (LSLU). We utilized the following benchmark datasets: ① Silkworm (20 classes of color images, 17k for training, and 6k for testing), ② CIFAR-10 [21](10 classes of color images, 50k for training, and 10k for testing), and ③ CIFAR-100 [21](100 classes of color images, 50k for training, and 10k for testing). Based on the characteristics of different networks, we devised two insertion methods: ① when employing the shallow network VanillaNet, replacing all custom activation functions with LSLU. ② when using other deep networks, employing LSLU only in selected downsampling layers. The reason for choosing method ② is that replacing all activation functions of a deep network during training resulted in inadequate fitting, as illustrated in Figure 3, which shows the outcome of replacing all activation functions in ResNet18[22] with LSLU.

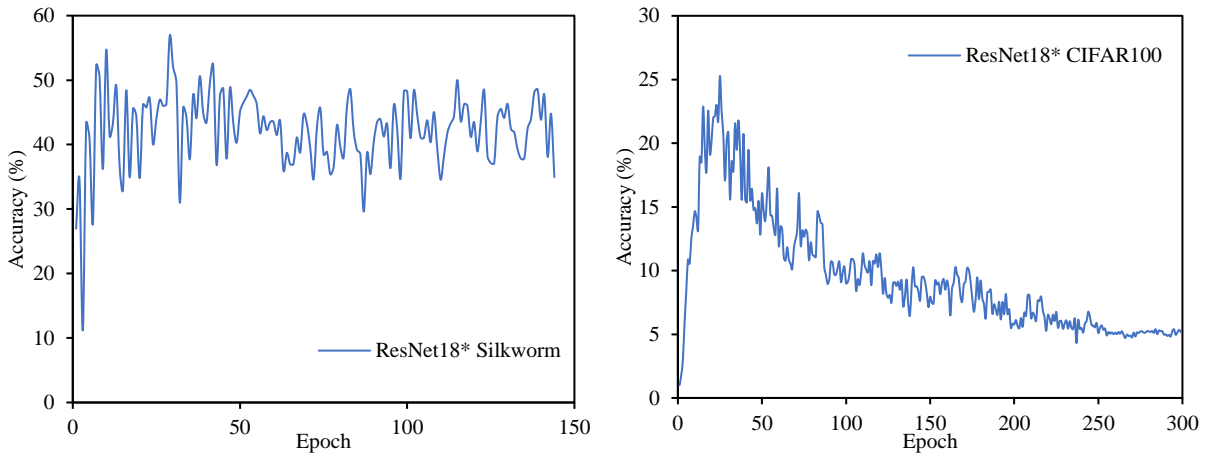

Figure 3: Incorrect use of LSLU. The specific settings for training parameters are detailed in Table 1.

*4.1 Experiment Settings*

According to the experimental task dataset and device parameters, our experimental design is as follows:

Table 1: CIFAR-100, CIFAR-10, and Silkworm training settings.

| Params | CIFAR-100 | CIFAR-10 | Silkworm |
|---|---|---|---|
| Base learning rate | VanillaNet: 3.5e-3(5,8,9,10)/4.8e-3(6,7) ResNet: 3.5e-3 | 3.5e-3 | 1e-3 |
| Learning rate schedule | Cosine decay[23] | Cosine decay | Cosine decay |
| EMA[24] | 0.999996 | 0.999996 | - |
| Early Stopping[25] | - | - | 7 |
| Optimizer | Lamb[26] | Lamb | Adam[27] |
| Loss Function | BCELoss[19] | BCELoss | CrossEntropyLoss[28] |
| GPU | Nvidia A800 | Nvidia 4060ti | Nvidia 4090 |
| Bach size | 800 | 128 | 16 |
| Epoch | 300 | 100 | - |

*4.2 CIFAR10 data set*

Let us validate the competitiveness of this nonlinearity relative to previous activation functions by borrowing experimental setups from Clevert et al[9] and Hendrycks et al[11]. For this purpose, we trained ResNet-18-LSLU ($n = 3$) on the CIFAR-10 dataset, adjusting it based on three different activation functions.

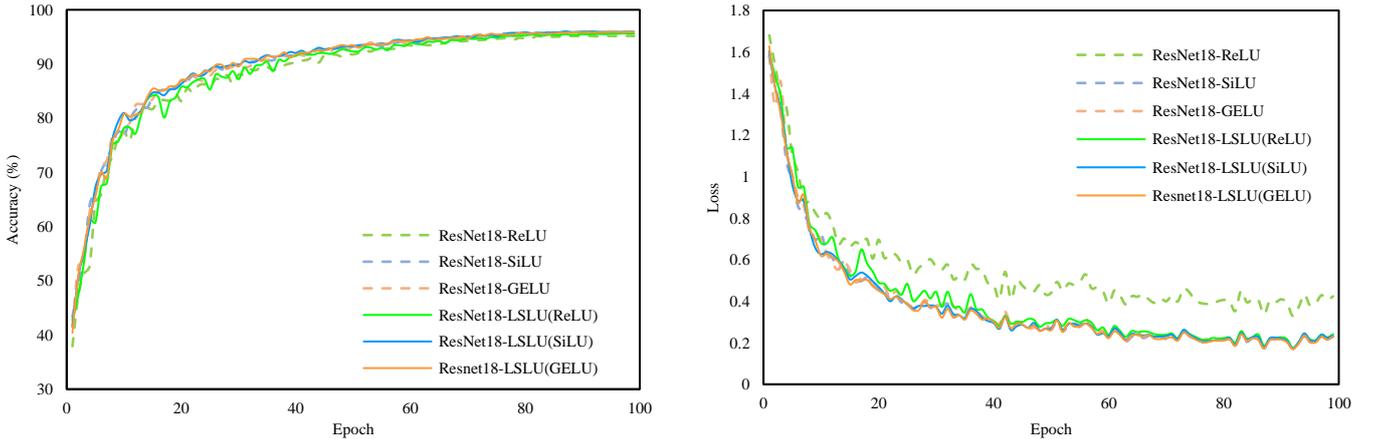

Figure 4: CIFAR10 Classification Results

**Cifar10 experimental results.** During the experiments, we uniformly used the Lamb optimizer with an initial learning rate ($\eta$) of $3.5 \times 10^{-3}$ and applied cosine annealing. The training was conducted for 100 epochs, with a batch size of 128. For LSLU, the parameters $\theta$ and $\omega$ were initialized to 1 and 0, respectively. The weight matrix and bias matrix were initialized with the identity matrix and zero matrix, respectively. As shown in the figure, after inserting LSLU, accuracy improved, and the loss was further reduced. Particularly, compared to the ReLU activation function, the difference in loss when each achieved its lowest was 0.1541. In Table 2, we provide detailed experimental result data. Each set of data underwent three experiments, and we report the maximum accuracy, average latency, and training speed, as well as the minimum loss. The latency test involved 10,000 images on Nvidia 4060ti, with a batch size of 1, measuring the time required for one inference. The final results demonstrate that the combination of LSLU with different activation functions yields varying improvements. When combined with SiLU and GELU, the best accuracy results were obtained, albeit with slower inference and training speeds.

Combining with ReLU yielded slightly lower accuracy but relatively faster inference speed.

Table 2: CIFAR-10 dataset experimental results

| Models | Params (M) | FLOPs (G) | | Top-1 Acc (%) | | Latency (ms) | | Trian seed (s/Epoch) | Min Loss |
|---|---|---|---|---|---|---|---|---|---|
| ResNet-18-ReLU[22] | 11.69 | 1.82 | | 95.21 | | **3.38** | | **103.91** | 0.329202 |
| ResNet-18-SiLU | 11.69 | 1.82 | | 95.9 | | 3.4 | | 140.4 | 0.171604 |
| ResNet-18-GELU | 11.69 | 1.82 | | 95.92 | | 3.39 | | 106.34 | 0.176683 |
| ResNet-18-LSLU(ReLU) | 11.69 | 1.86 | +0.04 | 95.67 | +0.46 | 3.49 | +0.11 | 135.83 | 0.175004 |
| ResNet-18-LSLU(SiLU) | 11.69 | 1.86 | +0.04 | **96.03** | +0.13 | 3.52 | +0.12 | 187.03 | 0.171918 |
| ResNet-18-LSLU(GELU) | 11.69 | 1.86 | +0.04 | 95.99 | +0.07 | 3.51 | +0.12 | 142.34 | **0.169825** |

**Analysis of parameter changes.** To understand the convergence of $\theta$ and $\omega$ during the network training process, we output and analyze them. For data rigor, we conducted two repeated experiments, and the final values were taken as the average of three values. As shown in Figure 5, it displays the convergence of $\theta$ and $\omega$ for the ResNet18 network using LSLU ($n$=3) with different activation functions. It is evident that the convergence curves for different activation functions are distinct. The $\theta$ curve for the ReLU activation function steadily increases and gradually stabilizes, while the SiLU and GELU activation functions exhibit an initial upward trend in the first few epochs, followed by a gradual decrease as training progresses until eventually stabilizing. As for $\omega$, the curves follow an unusual trend; the $\omega$ values for the ReLU activation function fluctuate around 0, while for SiLU and GELU, there is an initial upward trend in the first few epochs, and then $\omega_2$ and $\omega_3$ gradually decrease with training, while $\omega_1$ consistently increases but eventually stabilizes. According to Equations (15) and (16), $\theta$ and $\omega$ represent the slope and oscillation coefficient of the cascaded activation functions, and their updates follow the backpropagation algorithm. In Figure 5, we observe that both $\theta$ and $\omega$ tend to stabilize at a final value, and the convergence results are annotated in the graph. The convergence results represent the average of the last 10 epochs. Simultaneously, we calculated the standard deviation of the last 10 epochs, all of which are on the order of $10^{-3}$, indicating good stability.

In summary, both $\theta$ and $\omega$ iterate from initialization values to a stable state, but different activation functions exhibit distinct iteration trends. This interesting phenomenon suggests that the form of activation functions (slope and oscillation coefficient) should not remain constant during the training of neural networks. To further validate this hypothesis, we will continue experiments on the CIFAR100 dataset.

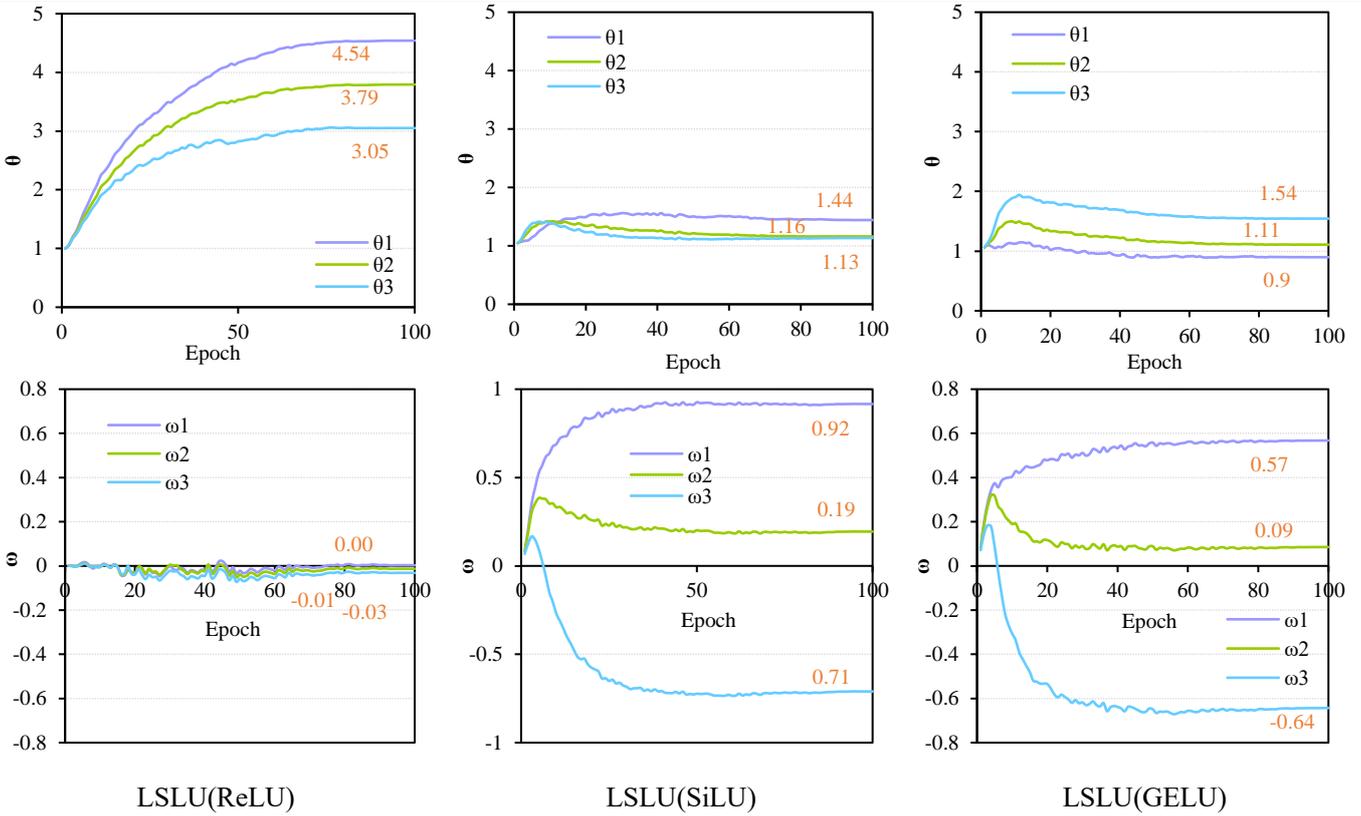

Figure 5: Illustrates the changes in learnable parameters. The *x*-axis represents the number of training epochs, while the *y*-axis depicts the updating process of learnable parameters $\theta$ and $\omega$. Both $\theta$ and $\omega$ undergo iterations from their initialization values to converge to a stable state.

*4.3 CIFAR100 data set*

To evaluate the performance of the Learnable Series Linear Unit (LSLU) and verify the convergence patterns of $\theta$ and $\omega$, we deployed it on VanillaNet and ResNet networks for experiments using the CIFAR-100 dataset. Due to LSLU's significantly superior single-layer nonlinearity, learning more parameters, we applied strong regularization. The latency test was conducted on Nvidia A800 GPU.

**Cifar100 experimental results.** We tested VanillaNet and ResNet architectures with varying numbers of layers. Table 3 presents the classification results on the CIFAR-100 dataset using different networks. We compared the number of parameters, FLOPs, depth, GPU latency, and accuracy. The architectures with the insertion of LSLU are appropriately marked in the table. To ensure the reliability of our experiments, we set the same hyperparameters for each type of architecture. The experiments revealed that after inserting LSLU (ReLU, $n=3$, dropout=0.1) into the VanillaNet series networks, the overall parameter count increased by 0.02-0.04M, FLOPs remained unchanged, latency increased by 0.04-0.11ms, and Top1-Acc improved by 1.28%-3.17%. For the ResNet series networks, the overall parameter count remained the same, FLOPs increased by 0.04G, latency increased by 0.26-0.64ms, and Top1-Acc improved by 0.02%-2.51%.

Table 3: Comparison of CIFAR-100. Latency is tested on Nvidia A800 GPU with batch size of 1.

| Network | Depth | Params(M) | | FLOPs(G) | | Latency(ms) | | Top1-Acc(%) | |
|---|---|---|---|---|---|---|---|---|---|
| VanillaNet-5[19] | 5 | 15.52 | | 5.16 | | 1.56 | | 76.98 | |
| VanillaNet-6[19] | 6 | 32.51 | | 5.99 | | 1.99 | | 78.12 | |
| VanillaNet-7[19] | 7 | 32.80 | | 6.89 | | 2.27 | | 79.40 | |
| VanillaNet-8[19] | 8 | 37.10 | | 7.74 | | 2.54 | | 79.52 | |
| VanillaNet-9[19] | 9 | 41.40 | | 8.58 | | 2.85 | | 79.96 | |
| VanillaNet-10[19] | 10 | 45.69 | | 9.42 | | 3.27 | | 81.59 | |
| VanillaNet-5-LSLU | 5 | 15.54 | +0.02 | 5.16 | | 1.67 | +0.11 | 78.65 | +1.67 |
| VanillaNet-6-LSLU | 6 | 32.53 | +0.02 | 5.99 | | 2.10 | +0.11 | 81.29 | +3.17 |
| VanillaNet-7-LSLU | 7 | 32.82 | +0.02 | 6.89 | | 2.23 | +0.04 | 82.48 | +3.08 |
| VanillaNet-8-LSLU | 8 | 37.13 | +0.03 | 7.74 | | 2.59 | +0.04 | 82.63 | +3.11 |
| VanillaNet-9-LSLU | 9 | 41.43 | +0.03 | 8.58 | | 2.91 | +0.06 | 82.73 | +2.77 |
| VanillaNet-10-LSLU | 10 | 45.73 | +0.04 | 9.42 | | 3.27 | | 82.87 | +1.28 |
| ResNet-18[22] | 18 | 11.69 | | 1.82 | | 3.16 | | 78.17 | |
| ResNet-34[22] | 34 | 21.80 | | 3.67 | | 5.96 | | 80.43 | |
| ResNet-50[22] | 50 | 25.56 | | 4.10 | | 7.84 | | 80.76 | |
| ResNet-101[22] | 101 | 44.55 | | 7.82 | | 11.85 | | 82.94 | |
| ResNet-18-LSLU | 18 | 11.69 | | 1.86 | +0.04 | 3.64 | +0.48 | 80.68 | +2.51 |
| ResNet-34-LSLU | 34 | 21.80 | | 3.71 | +0.04 | 6.60 | +0.64 | 81.14 | +0.71 |
| ResNet-50-LSLU | 50 | 25.56 | | 4.14 | +0.04 | 8.45 | +0.61 | 81.23 | +0.47 |
| ResNet-101-LSLU | 101 | 44.55 | | 7.86 | +0.04 | 12.11 | +0.26 | 82.96 | +0.02 |

**Results analysis.** The significant improvement in accuracy for VanillaNet and ResNet after using LSLU indicates the effectiveness of the LSLU design. The VanillaNet-LSLU series networks outperform the original networks in accuracy, suggesting that combining shallow networks with high nonlinearity-capable activation functions has broad prospects in image recognition. The ResNet-LSLU family of networks achieve higher accuracy compared to the original network. This suggests that using activation functions with high non-linearity after certain downsampling layers in deep networks is necessary.

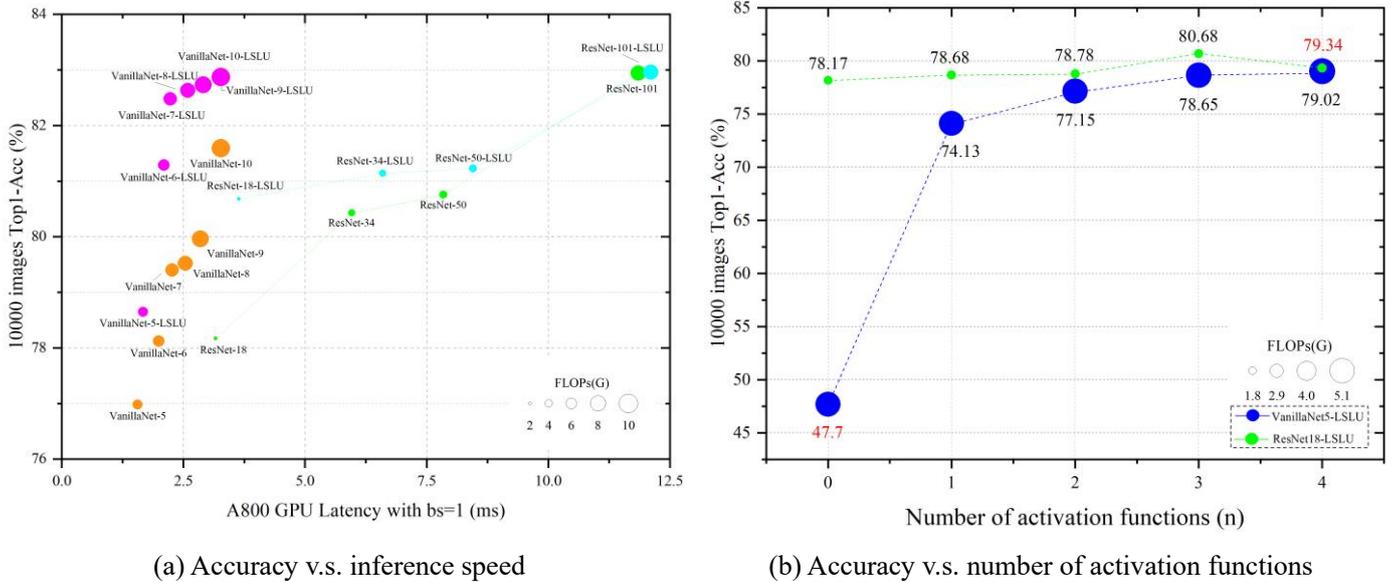

(a) Accuracy v.s. inference speed  (b) Accuracy v.s. number of activation functions

Figure 6: Top-1 Accuracy on CIFAR-100 vs. Inference Speed on Nvidia A800 GPU with batch size 1. The size of the circle is related to the value of FLOPs, and different colors represent different network architectures. In (a), the horizontal axis

represents the inference speed, and the vertical axis represents the Top-1 accuracy on the test set. In (b), the horizontal axis represents the number of activation functions used, and the vertical axis represents the Top-1 accuracy on the test set. Figure 6(a) displays the accuracy and inference speed for different architectures. The inference speed at batch size 1 is highly correlated with network depth and independent of the number of parameters, indicating significant potential for real-time processing in simple shallow networks. It can be observed that inserting LSLU into these architectures slightly increases GPU latency but achieves better accuracy. This suggests that inserting LSLU has advantages when there is sufficient computational power.

**Analysis of the impact of the number of activation functions.** Additionally, we conducted experiments to investigate the impact of the number of activation functions $n(0,1,2,3,4)$ on the model. When $n = 0$, only regular ReLU activation functions are used. The experimental results are shown in Figure 6(b). It is easy to see that as $n$ increases, the overall accuracy of the model tends to be higher. Upon careful observation of Figure 5(b), it is surprising to find that there is an anomaly in accuracy when $n = 0, 4$. Table 4 presents the specific test values, and through analysis, two interesting conclusions can be drawn: ① VanillaNet-5-LSLU is a shallow and wide network that requires high nonlinearity for a single layer. When $n = 0$, regular activation functions cannot provide such high nonlinearity. ② ResNet-18-LSLU is a deep and narrow network, and its depth determines that a single-layer nonlinearity does not need to be too high. Therefore, when $n = 4$, excessive nonlinearity leads to a decrease in accuracy.

Table 4: Experimental results of ResNet18-LSLU and VanillaNet5-LSLU on CIFAR-100 when the number of activation functions is varied.

| Network | n | Params (M) | FLOPs (G) | Top-1 Acc (%) | Latency (ms) |
|---|---|---|---|---|---|
| ResNet-18-LSLU | 0 | 11.69 | 1.82 | 78.17 | 3.55 |
|  | 1 | 11.69 | 1.83 | 78.68 | 3.58 |
|  | 2 | 11.69 | 1.84 | 78.78 | 3.63 |
|  | 3 | 11.69 | 1.86 | 80.68 | 3.64 |
|  | 4 | 11.69 | 1.89 | 79.34 | 3.97 |
| VanillaNet-5-LSLU | 0 | 15.17 | 5.02 | 47.7 | 1.48 |
|  | 1 | 15.23 | 5.04 | 74.13 | 1.53 |
|  | 2 | 15.35 | 5.09 | 77.15 | 1.54 |
|  | 3 | 15.54 | 5.16 | 78.65 | 1.67 |
|  | 4 | 15.78 | 5.26 | 79.02 | 1.72 |

**Different numbers of activation functions.** In Figure 5, we observed the trends of $\theta$ and $\omega$ in ResNet-18-LSLU with different activation functions and inferred that the form of activation functions (slope and oscillation coefficient) should not remain constant during the training of neural networks. To validate this hypothesis, we outputted the trends of $\theta$ and $\omega$ in the Stages.0.act layer of the VanillaNet5 network using different numbers of activation functions. As shown in Figure 7, it is evident that when $n=1,2,3,4$, $\theta$ initially drops suddenly, then gradually rises and eventually stabilizes. $\omega$, on the other hand, slowly decreases and eventually stabilizes. Combined with Table 4, when $n=0$, i.e., using only static ReLU, the accuracy of ResNet-18 is 78.17%, and VanillaNet5 is 47.7%, both much lower than the accuracy achieved with LSLU.

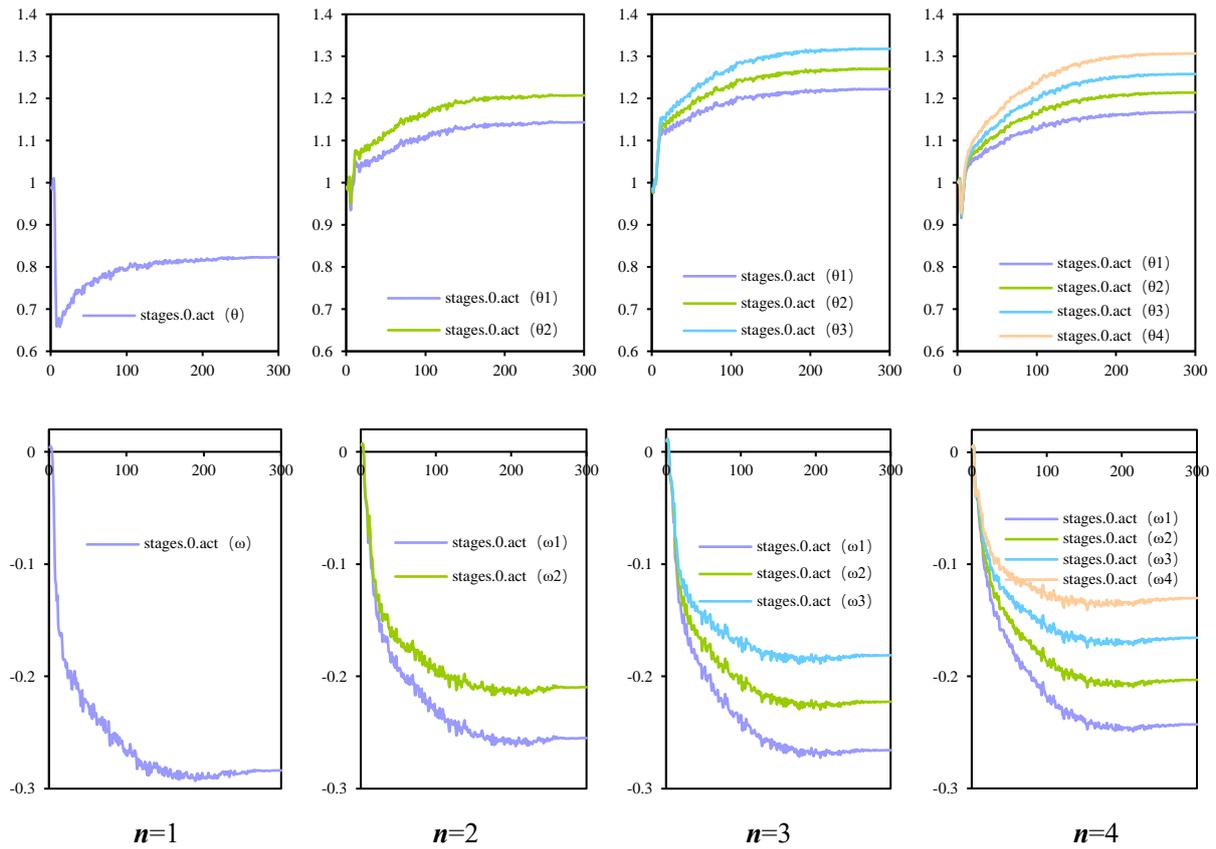

Figure 7: Shows the trends of $\theta$ and $\omega$ with different values of $n$. The $x$-axis represents the number of training epochs, and the $y$-axis represents $\theta$ or $\omega$. $\theta$ initially experiences a sudden drop, then gradually rises, eventually stabilizing. On the other hand, $\omega$ exhibits a slow descent, gradually stabilizing over time.

**Different depth.** To further validate the hypothesis, we decided to analyze the convergence patterns of $\theta$ and $\omega$ when using VanillaNet networks with different depths. As shown in Figure 8, it demonstrates the trends of $\theta$ and $\omega$ in the Stages.4.act layer of VanillaNet-LSLU networks with depths ranging from 7 to 10 layers. From the graph, it can be observed that the trends of $\theta$ and $\omega$ are similar across networks with different depths but with slight variations. As the network depth increases, the gaps between $\theta$ ($\theta_1, \theta_2, \theta_3$) and $\omega$ ($\omega_1, \omega_2, \omega_3$) values gradually narrow. Moreover, $\theta$ overall tends to approach the initial value of 1, and $\omega$ gradually moves away from the $x$-axis. In general, in conjunction with Table 3, it can be noted that the accuracy of each network increases with the growth of network depth, consistent with the trends in $\theta$ and $\omega$.

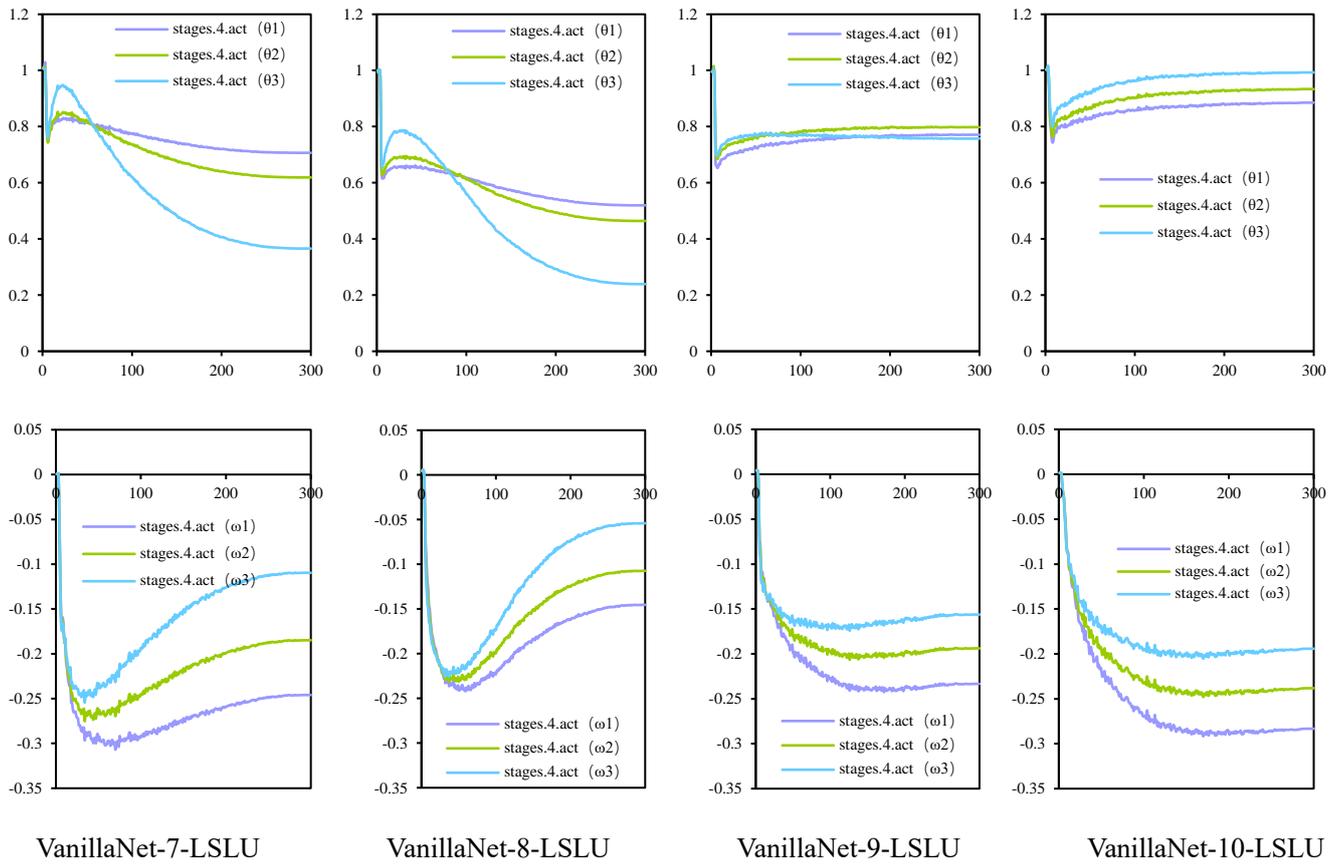

VanillaNet-7-LSLU    VanillaNet-8-LSLU    VanillaNet-9-LSLU    VanillaNet-10-LSLU

Figure 8: Network Depth Impact. The trends of $\theta$ and $\omega$ in networks with different depths are generally similar but with slight variations. As the network depth increases, the gaps between $\theta$ ($\theta_1$, $\theta_2$, $\theta_3$) and $\omega$ ($\omega_1$, $\omega_2$, $\omega_3$) values gradually narrow. Moreover, $\theta$ overall tends to approach the initial value of 1, and $\omega$ gradually moves away from the $x$-axis.

**The impact of regularization strength.** We found that the use of regularization with different strengths has an impact on the network models. As shown in Figure 9, as the regularization strength increases, the validation accuracy of the model first improves and then deteriorates, while the loss gradually decreases. However, both accuracy and loss are still better than when LSLU is not used. In general, the accuracy of VanillaNet-6 and VanillaNet-5 networks improves with a slight increase in regularization[29]. VanillaNet-5 and VanillaNet-6 networks use different dropout rates ( $p = 0.0, 0.1, 0.2, 0.3, 0.4$ ), resulting in a convex curve for the final accuracy results. When the dropout rate is around 0.2, the network accuracy approaches the optimal value.

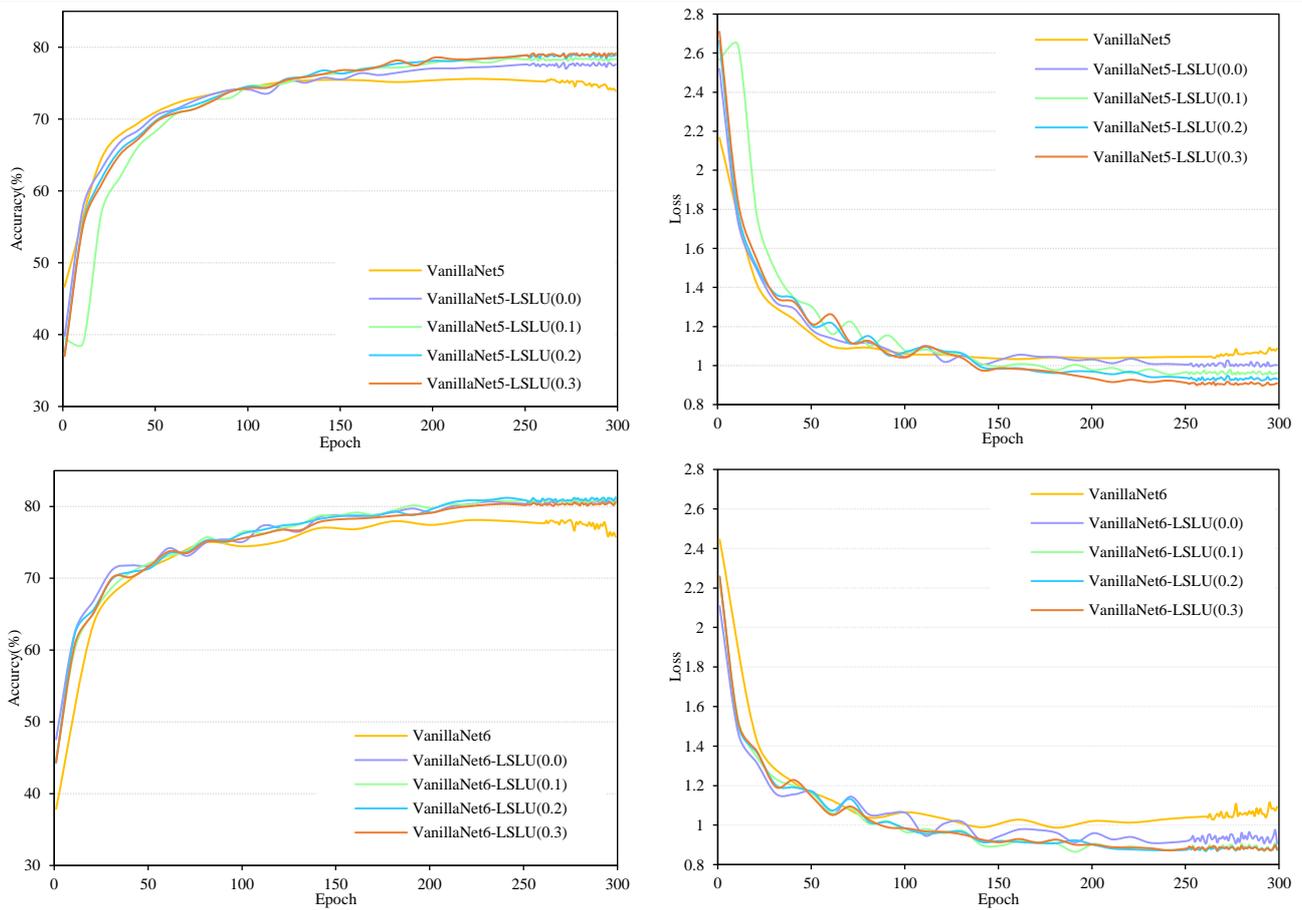

(a)

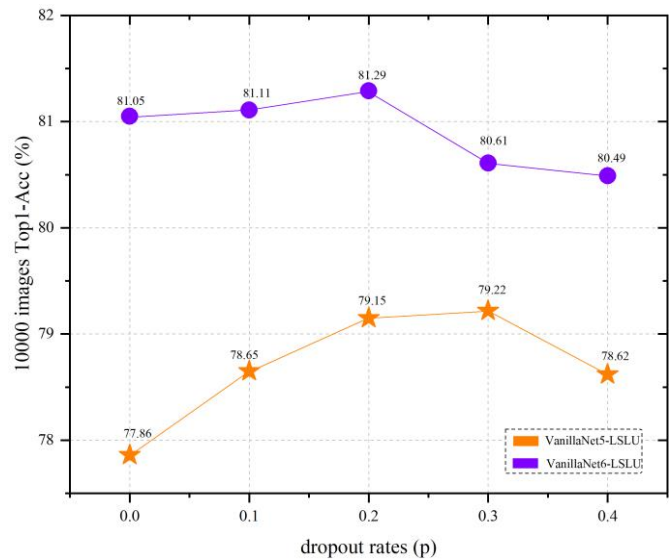

(b)

Figure 9: Different strengths of regularization. (a) VanillaNet-6 and VanillaNet-5 classification accuracy and loss curve. (b) VanillaNet-5-LSLU and VanillaNet-6-LSLU classification results with different dropout rate values. They use the same training strategy, with the only difference being the strength of regularization.

**Class selectivity index distribution.** The network utilizing LSLU exhibits distinctive feature characteristics. To understand this, we conducted class selectivity index analysis on the training weights of VanillaNet-5 and ResNet-18. The class selectivity index measures the difference between the highest class-conditioned average activity and the average activity conditioned on all other classes. The final normalized value ranges between 0 and 1, where 1 indicates that the filter is activated only for a single class, and 0 indicates uniform activation across all classes[30]. In Figure 10, we plot the class selectivity index distributions for all layers in VanillaNet-5 and ResNet-18 models. Overall, the orange and green areas mostly overlap, which is what we hope to see, indicating that the feature selection for classification is largely consistent after using both activation methods. Where they do not overlap, it suggests that the different activation methods result in varying trends in classification selection. In the shallow stages, the VanillaNet-5 curve shows a phenomenon of high and narrow peaks, indicating that the feature learning in the network produces some focusing effects, making regions with relatively higher selectivity indices more concentrated, such as Stages Block0 Conv0. At the same time, the use of LSLU enhances the response to specific categories. However, as the layers become deeper, they start to diverge, meaning that in these layers, the selectivity indices for different categories are relatively evenly distributed compared to the original network. For example, Cls1 and Cls2, the use of LSLU leads to more dispersed and blurred features. Whether in shallow or deep layers, if the green area exceeds the orange area, it indicates that the peak values and widths of the curve after using LSLU surpass those of the original network, also suggesting that after using LSLU, the network's feature selection range for classification is greater than before. In summary, the figure shows that with the use of LSLU (both peaks and widths), the network tends to include more class-specific features than the original network.

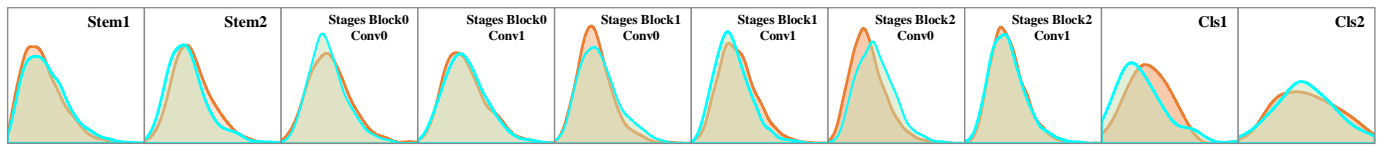

VanillaNet-5 v.s VanillaNet-5-LSLU

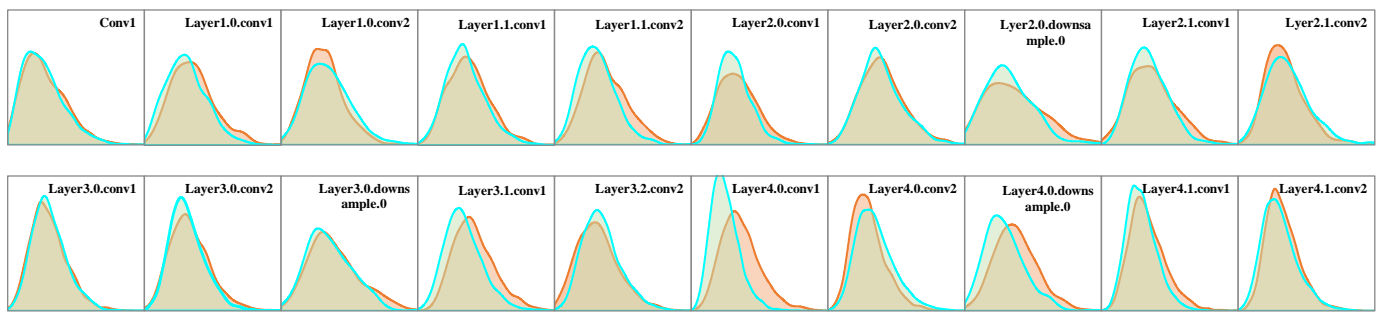

ResNet-18 v.s ResNet-18-LSLU

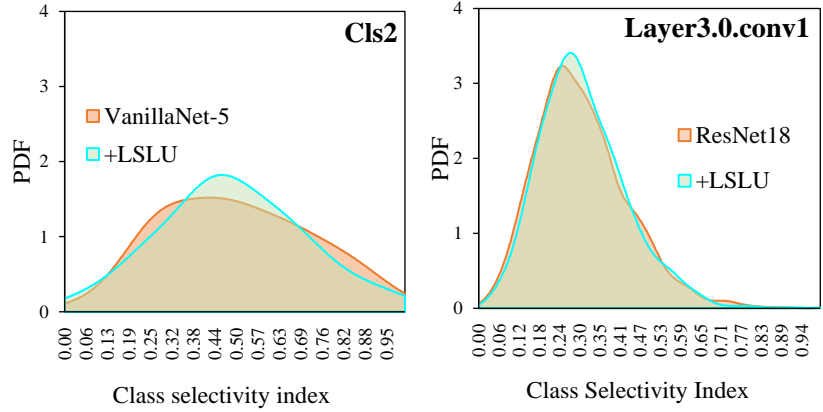

Figure 10: Class selectivity distribution. The *x*-axis and *y*-axis show the class selectivity index and its density (PDF), respectively. Using the CIFAR-100 validation dataset, we calculate the exponential distribution of class selectivity between the original network (orange) and the network after using LSLU (green). We can easily see that with LSLU, the network will tend to include more class-generic features.

*4.3 Silkworm data set*

Whether it is the neural network architecture or auxiliary enhancement modules, their design, while relying on common datasets to demonstrate their effectiveness, still needs adjustments according to the specific task of the dataset during application. In this subsection, we use LSLU in various neural networks to learn the Silkworm classification dataset, comparing the benefits before and after using LSLU to analyze its performance on a specific task.

**Comparative analysis of multiple models.** As shown in Figure 11, the training processes on ResNet18[22], MobileNetV3-L[31], ShuflleNetV2-1×[32], EfficientNetV2-S[33], and VanillaNet-5[19] are displayed. It is evident that after inserting LSLU, the accuracy learning curve reaches higher levels, and the loss curve decreases further.

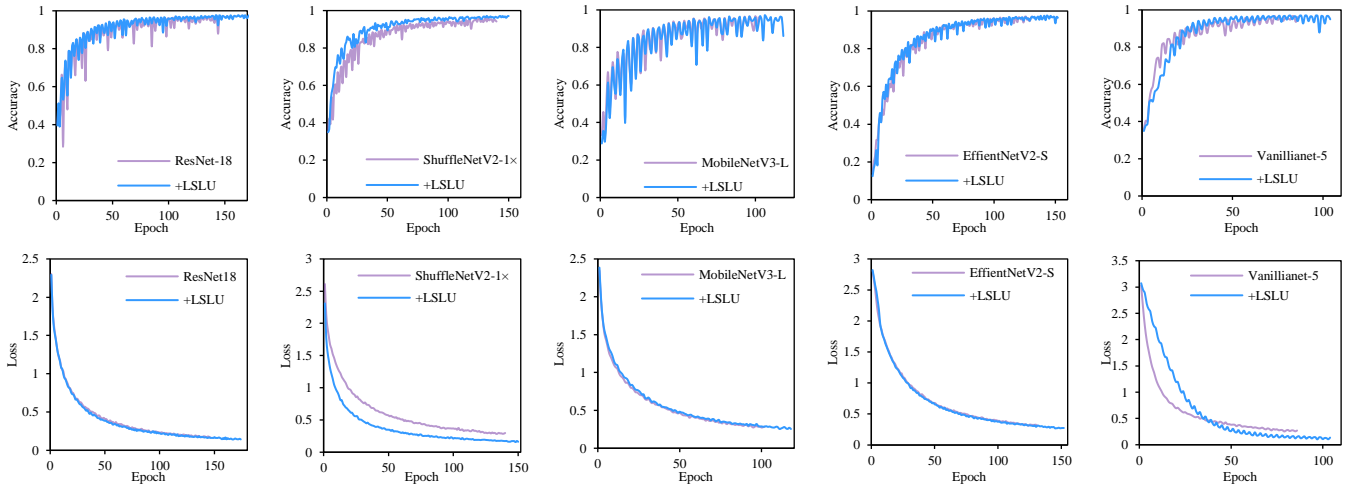

Figure 11: Silkworm classification results. When using LSLU, the accuracy learning curves of various networks rise higher, and the loss curves decrease lower.

The final experimental results are shown in Table 5, and it is easy to see that the recognition accuracy of each network has significantly improved with the use of LSLU.

Table 5: Comparison of Silkworm. Latency is tested on Nvidia 4090 GPU with batch size of 1. It is easy to observe that with the use of LSLU, despite a slight increase in latency, there is a significant improvement in accuracy.

| Network | Params(M) | FLOPs(G) | Latency(ms) | Top1-Acc（%） |
|---|---|---|---|---|
| ResNet-18 | 11.69 | 1.82 | 9.17 | 96.86 |
| ResNet-18-LSLU | 11.69 | 1.86 | 10.25 | 97.33 |
| ShuffleNetV2-1× | 1.27 | 0.15 | 17.88 | 95.96 |
| ShuffleNetV2-1×-LSLU | 2.28 | 0.16 | 21.07 | 97.18 |
| MobileNetV3-L | 5.48 | 0.22 | 21.67 | 96.71 |
| MobileNetV3-L-LSLU | 5.48 | 0.23 | 22.56 | 97.07 |
| EfficientNetV2-S | 20.19 | 2.86 | 63.60 | 97.07 |
| EfficientNetV2-S-LSLU | 20.19 | 2.88 | 62.72 | 97.51 |
| VanillaNet-5 | 15.52 | 5.16 | 2.46 | 96.59 |
| VanillaNet5-LSLU | 15.54 | 5.16 | 2.69 | 96.86 |

*4.4 Adaptive Correction Theory*

From the overall performance on the CIFAR-10 and CIFAR-100 datasets, it is evident that regardless of the number of basic activation functions or the depth of the network, there always exists a worst-case activation function determined by the learnable parameters $\theta$ and $\omega$. As training progresses, the activation function is gradually corrected, and this correction process has no limit. These experimental results bear some similarity to the study by Sun et al.[34]: The worst activation function with boundary conditions (WAFBC) can be optimized using the EAFO method based on information entropy (the classification effect is better when the probability distributions of the two classes differ significantly, i.e., when the information entropy is low), deriving a new activation function from the existing ones. Therefore, it can be concluded that the LSLU method adaptively corrects the original activation function $f(x)$ using a suffix term, thereby obtaining a more suitable activation function for that activation layer.

*4.5 Modification Suggestions*

Therefore, we recommend using LSLU judiciously based on information such as the network's structure and depth: for shallow networks like VanillaNet, LSLU can be fully substituted for the old activation functions. However, for deep networks such as ResNet, and EfficientNetV2, which inherently possess high nonlinearity, careful consideration is required when selecting the position, number of activation functions, and the values of dropout rates for LSLU usage.

# 5 Conclusion

This study proposes a Learnable Series of Linear Units (LSLU) designed for convolutional neural network structures to enhance the nonlinearity of activation function layers. Validation experiments were conducted on three datasets: CIFAR-10,

CIFAR-100, and Silkworm. On CIFAR-10, experiments were conducted with different activation functions combined with LSLU. The results showed that the SiLU and GELU achieved the best accuracy but at the cost of lower inference and training speeds. Combining with ReLU resulted in slightly lower accuracy but relatively faster inference speed. On CIFAR-100, experiments were performed with both shallow and deep networks combined with LSLU. It was observed that replacing all activation functions in deep networks leads to overfitting issues, but using LSLU only in certain downsampling layers demonstrates the effectiveness of improvement. For shallow networks, replacing with LSLU increased accuracy. On the Silkworm dataset, LSLU was validated in specific applications, showing improved accuracy for various network architectures. At the end of the article, we provide recommendations for using LSLU. Learnable Series Linear Units are still in the early stages of development, and their application and research play a crucial role in addressing the interpretability issues in neural networks.

**Limitations and Future Work.** Our research results raise several important questions for future work. First, the LSLU method cannot replace all activation function layers simultaneously in networks other than VanillaNet, and extensive experiments are required to determine the number and distribution of activation functions. This leads to additional training costs, and future work needs to explore how to use the LSLU method more conveniently to minimize training costs. Secondly, an exciting direction for future research is to investigate the mathematical reasons behind the variation of $\omega$ in the learnable parameter $\theta$ when using the LSLU method. Finally, although we have validated the feasibility of the LSLU method in image classification tasks, its performance in other tasks remains to be explored.

# CRediT authorship contribution statement

**Chuan Feng:** Writing - original draft, Writing - review & editing. **Shiping Zhu:** Conceptualization, Methodology, Resources. **Hongkang Shi:** Validation. **Xi Lin**: Data curation. **Maojie Tang** and **Hua Huang:** Formal analysis.

# Declaration of Competing Interest

The authors declare that they have no known competing financial interests or personal relationships that could have appeared to influence the work reported in this paper.

# Acknowledgments

The authors appreciate the silkworm samples provided by the Sericulture Research Institute of Sichuan Academy of Agricultural Sciences. This work is supported by the General Program of the National Natural Science Foundation of China: Research on the Mechanism of Mature Silkworm Cocooning and Online Detection Method for Cocoon Quality Integrating Visual and Spectral Features (32471991), the Ministry of Education's Industry-University-Research Innovation Fund—Dezhou Special Project: Research on Deep Learning-Based Automatic Grading Algorithm for Tobacco Leaves (2021DZ005), and the General Program of the Sichuan Provincial Natural Science Foundation: Research on Early Detection of Silkworm Pebrine Disease Based on Deep Learning and Behavior Recognition (2023NSFSC0498).